\documentclass[a4paper,conference]{IEEEtran}

\IEEEoverridecommandlockouts

\usepackage{cite}
\usepackage{graphicx}
\usepackage{float}
\usepackage{amsmath,amssymb,amsfonts}
\usepackage{bm}
\usepackage{mathtools}
\usepackage{booktabs}
\usepackage{multirow}
\usepackage{tabularx}

\usepackage{mathrsfs}
\usepackage[hidelinks]{hyperref}

\usepackage{graphicx}
\usepackage{subcaption} % for subfigure environments
\usepackage[table,dvipsnames]{xcolor}
\usepackage[dvipsnames]{xcolor} % dvipsnames enables ForestGreen, etc.
\usepackage{makecell}

\usepackage{xstring}

\title{GuidaPA: Privacy-Preserving Chatbot for Public Administration via Federated Learning}

\author{
\IEEEauthorblockN{
Daniel M. Jimenez-Gutierrez\textsuperscript{1},
Albenzio Cirillo\textsuperscript{2},
Raffaele Nicolussi\textsuperscript{2},
Alessio Beltrame\textsuperscript{2},
Andrea Vitaletti\textsuperscript{1}
}
\IEEEauthorblockA{\textsuperscript{1}Sapienza University of Rome, Via Ariosto 25, 00185 Rome, Italy}
\IEEEauthorblockA{\textsuperscript{2}Fondazione Ugo Bordoni, Viale del Policlinico 147, 00161 Rome, Italy}
\IEEEauthorblockA{Emails: jimenezgutierrez@diag.uniroma1.it, albenzio.cirillo@gmail.com,\\
rnicolussi@fub.it, a.beltra@gmail.com,vitaletti@diag.uniroma1.it}
}

\begin{document}
\maketitle

\begin{abstract}

We present \emph{GuidaPA}, a privacy-preserving chatbot for the Italian Public Administration (PA) trained via Federated Learning (FL) on documentation from two national PA platforms, SIGESON and SIDFORS. Our corpus includes approximately 8 pages of SIGESON manuals and 31 pages of SIDFORS manuals/FAQs; while this study uses public documentation as a safe proxy, the intended deployment extends to restricted internal sources (e.g., tickets, officer manuals, database extracts) that can not be centrally pooled due to regulatory and organizational constraints. GuidaPA integrates role-based access control, secure client-side preprocessing, explicit monitoring of non-IID effects, and parameter-efficient federated fine-tuning of large language models. Using QLoRA (4-bit) over 15 federated rounds with an 80/20 train--test split per client, we evaluate answer quality with ROUGE, BLEU-4, and METEOR. The best federated model achieves ROUGE-1/2/L of 61.10/55.77/59.44, BLEU-4 of 45.02, and METEOR of 63.94—close to private centralized fine-tuning while keeping data on-site. Compared to the general-purpose baseline, domain fine-tuning improves ROUGE-1 from 41.45 to 62.18 and BLEU-4 from 26.97 to 50.90. Overall, the results indicate that FL can deliver high-quality conversational AI for public services without centralized data sharing.

\end{abstract}


%%%%%%%%%%%%%%%%%%%%%%%%%%%%%%%%%%%%%%%%%
%%%%%%%%%%%%%%%%%%%%%%%%%%%%%%%%%%%%%%%%%
\section{Introduction} \label{sec:intro}

The rapid diffusion of Large Language Models (LLMs) is enabling wider adoption of NLP components based on Generative AI (GenAI) within public services~\cite{NLPDGDIGIT,agid_piano_triennale_2024}. Public Administration (PA)—the set of governmental institutions responsible for delivering public services and enforcing regulations~\cite{eu_interoperability_framework}—can benefit from LLM-based assistants to automate repetitive requests and improve access to information~\cite{oyewole2024automating}. However, PA constraints make conventional centralized training impractical and often undesirable from a privacy and governance perspective~\cite{tseng2023understanding,gokccearslan2024benefits}.

\begin{figure}[t]
    \centering
    \includegraphics[width=\linewidth]{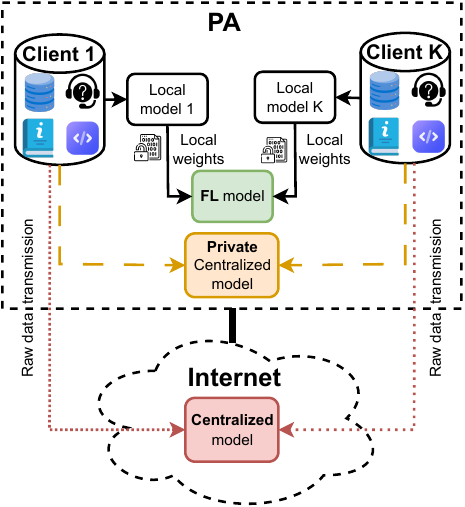}
    \caption{Alternative architectures for integrating an LLM within PA. Internet-centralized approaches transfer data to third parties; private centralized approaches keep data within PA but still require inter-entity sharing; Federated Learning (FL) exchanges only model updates, preserving data sovereignty.}
    \label{fig:three_architectures}
\end{figure}

Figure~\ref{fig:three_architectures} summarizes three deployment options. Internet-centralized solutions are technically convenient but require sending sensitive information to external providers, introducing privacy risks. A Private Centralized model keeps data within PA infrastructure and aligns with the principle that PAs should make data available to other PAs when required for institutional tasks, under legal and privacy constraints~\cite{CAD}. A centralized architecture within PA serves as a unified knowledge aggregator, effectively mitigating the inconsistencies generated by siloed workflows and ensuring seamless interoperability across all administrative sectors.
Nonetheless, inter-office pooling still increases governance burden and exposure. FL strengthens privacy by keeping data local and sharing only model updates, enabling collaborative learning across institutional silos while preserving data sovereignty.

\subsection{Motivation} \label{sec:motiv}
Deploying chatbots in PA is challenging due to organizational and regulatory constraints~\cite{chen2023adoption,dreyling2024challenges}. In our case study, these constraints arise in two operational PA platforms: SIDFORS and SIGESON. Their documentation includes end-user manuals, FAQs, and procedural guidance for administrative workflows such as enrollment, authorization requests, access to company accounting positions, annual declarations, and payment of administrative fees. Although this study uses only public documentation as a safe proxy, these sources reflect the structured and procedural nature of PA knowledge and represent the broader institutional material that a deployed system may need to support, including internal manuals, support tickets, and curated database extracts.

Key challenges include: (i) protecting sensitive data ($C_1$), (ii) adapting foundation models to PA-specific procedures and terminology ($C_2$), and (iii) integration with heterogeneous systems and operational scalability ($C_3$). Fine-tuning and Retrieval-Augmented Generation (RAG) can improve domain knowledge~\cite{lin2024data,lewis2020retrieval}, but they do not remove the need for an infrastructure that preserves privacy and supports distributed governance. As noted by the Italian Ministry of Economy and Finance’s Treasury Department (MEF--DT)~\cite{mef_dt_nota1_2024}, AI in PA requires robust infrastructure and data management; yet interoperability remains incomplete~\cite{bellitti2023pdnd}, making centralized data pooling unrealistic. Fully local \emph{ad hoc} solutions are possible but costly and fragmented, potentially leading to inconsistent behavior across offices.

\subsection{Research Objectives}
This study investigates whether FL can enable effective, privacy-preserving LLM-based chatbots in PA settings where data are distributed across institutional silos and cannot be centrally aggregated. We pursue: (i) designing a federated architecture with role-based access control (RBAC) and secure client-side data handling, (ii) evaluating whether federated fine-tuning matches private centralized quality while preserving locality, and (iii) assessing robustness under client heterogeneity and quantifying non-IID effects on global and local performance~\cite{jimenez2024non,ma2022state}.

\subsection{Contribution} \label{sec:contrib}
We introduce \emph{GuidaPA}, a privacy-preserving PA chatbot trained via FL to address $C_1$--$C_3$. Unlike prior FL applications that mainly focus on privacy-preserving model training alone, GuidaPA combines federated LLM adaptation with PA-specific deployment mechanisms that are directly relevant in practice: role-based access control, secure client-side preprocessing, and explicit monitoring of non-IID effects. Its main technical contribution, therefore, goes beyond applying FL to chatbot adaptation by integrating these components into a practical framework for institutional settings and showing empirically that this design can achieve near-centralized quality without centralizing the underlying documents. Using real documentation from the SIDFORS and SIGESON platforms, federated fine-tuning achieves quality comparable to a private centralized approach (Section~\ref{baseline}) while keeping data local, and consistently improves over the general-purpose baseline (Table~\ref{tab:metric-centralized}), demonstrating effective domain adaptation. Since PA documentation follows shared standards and regulatory structures, cross-client non-independent and identically distributed (non-IID) data is limited but non-zero, allowing federated training to converge while capturing realistic non-IID effects~\cite{jimenez2024non,ma2022state}. More broadly, this work shows how privacy-preserving learning architectures can support controlled dissemination and access to institutional knowledge in public-sector information systems.

\subsection{Paper Structure}
Section~\ref{sec:related-work} reviews related work. Section~\ref{sec:data-sources} describes the data sources and dataset. Section~\ref{sec:system} presents the architecture. Sections~\ref{sec:experiments} and~\ref{sec:results} report the experimental setup and results, followed by discussion (Section~\ref{sec:discussion}), limitations (Section~\ref{sec:limitations}), and conclusions (Section~\ref{sec:conclusion}).

%%%%%%%%%%%%%%%%%%%%%%%%%%%%%%%%%%%%%%%%%
%%%%%%%%%%%%%%%%%%%%%%%%%%%%%%%%%%%%%%%%%
\section{Related Work}
\label{sec:related-work}

The use of chatbots in the context of PA remains relatively limited compared to commercial and open-domain applications, where large-scale conversational systems benefit from abundant centralized data and rapid iteration cycles. To the best of our knowledge, no prior work has applied FL to the development of LLM--based chatbots in the PA domain in Italy. Prior research spans (i) centralized chatbots for e-government and public services, (ii) federated chatbot approaches in privacy-sensitive domains, and (iii) federated fine-tuning of LLMs, which provides methodological foundations for privacy-preserving adaptation.

Centralized PA chatbots have been proposed for citizen support and information access, including Italian public-service conversational frameworks and domain assistants~\cite{piizzi2024natural,bellini2020guapp}, as well as systems for navigating governmental open data and improving transparency and responsiveness~\cite{cantador2021chatbot,androutsopoulou2019transforming}. While these works demonstrate the potential of conversational interfaces to reduce friction in citizen--government interactions, they typically assume that training data can be collected and maintained in a single location, which is often unrealistic in PA environments characterized by organizational silos and constrained data governance.

In privacy-sensitive domains such as healthcare and education, FL has been used to train conversational agents without sharing raw conversations, showing that collaborative learning is feasible under data-locality constraints~\cite{ait2023fedbot,puppala2024scan,d2023boulez}. However, these studies do not address PA-specific requirements such as role-based access control, institutional governance, and cross-office procedural heterogeneity. Our work builds on this literature by focusing on these deployment constraints in the PA setting.

Recent surveys and methods on federated LLM adaptation highlight key challenges and enablers---including communication efficiency, non-IID data, and parameter-efficient fine-tuning (PEFT) with low-rank adaptation and quantization---that motivate our design choices~\cite{yao2024federated,wang2024flora}. PEFT reduces the computational and communication cost of federated tuning, while raising questions about stability and aggregation under client heterogeneity. These issues are especially relevant in PA, where local datasets may be small but authoritative, and compute resources are limited.

GuidaPA addresses this gap by combining federated fine-tuning of LLMs with an application-level architecture tailored to PA requirements, including RBAC, client-side secure preprocessing, and explicit monitoring of non-IID data effects. In doing so, this work bridges recent advances in federated LLM research with the practical needs of public-sector information services, providing empirical evidence that near-centralized quality can be achieved without centralizing the underlying documents.

% %%%%%%%%%%%%%%%%%%%%%%%%%%%%%%%%%%%%%%%%%
% %%%%%%%%%%%%%%%%%%%%%%%%%%%%%%%%%%%%%%%%%
\section{Domain-Specific Data Sources and Dataset Description for GuidaPA}
\label{sec:data-sources}

GuidaPA was trained using domain documentation from two operational systems managed by the Ministry of Enterprises and Made in Italy (MIMIT) and developed with the support of Fondazione Ugo Bordoni (FUB), namely SIDFORS (\emph{Sistema per dichiarazione fornitura di reti/servizi di comunicazione elettronica})~\cite{SIDFORS} and SIGESON (\emph{Sistema di Gestione delle Reti di Radiodiffusione sonora})~\cite{SIGESON}. We use their publicly available manuals and FAQs as training and evaluation sources. Public docs are used here as a safe proxy for restricted internal sources.

\textbf{SIDFORS} enables sector operators to submit general authorization requests for electronic communications networks and services (e.g., telephony, Wi-Fi, satellite, fiber), request authorization to access a company’s accounting position, and pay the applicable fees~\cite{SIDFORS}.

\textbf{SIGESON} supports audio broadcasting operators in requesting authorization to access a company’s accounting position and paying the applicable fees~\cite{SIGESON}.

In both cases, requests submitted by operators must undergo manual review and validation by public officers, and users often contact officers to clarify procedures and platform functionality. To reduce support effort, both platforms provide freely accessible documentation; for this study, GuidaPA is trained exclusively on this public material:
\begin{itemize}
\item End-user manuals describing core procedures (e.g., enrollment, authorizations, payments, access to accounting positions)~\cite{SIDFORS,SIGESON}.
\item FAQs addressing common issues encountered by end-users~\cite{SIDFORS,SIGESON}.
\end{itemize}

\subsection{Dataset Composition and Statistics}
The datasets consist of Italian domain documentation from two PA platforms, each corresponding to one federated client and reflecting a distinct administrative vertical.

\emph{Client 1: SIGESON (Radio Broadcasting).}
The SIGESON dataset is derived from the user manual ``Guida rapida per l'operatore'' (Version 1, October 2023; $\sim$8 pages) and the platform FAQ knowledge base~\cite{SIGESON}. It focuses on workflows such as access authorization to the accounting module, annual turnover declarations, and payment of administrative dues.

\emph{Client 2: SIDFORS (Electronic Communications).}
The SIDFORS dataset includes the ``Manuale d’uso del servizio SIDFORS'' (Version 2.1, May 2024; $\sim$31 pages) and the platform FAQs~\cite{SIDFORS}. It covers general authorization workflows (e.g., fiber, satellite, MNO/MVNO) and procedures to access the accounting module for managing pagoPA-related payments.

\textbf{Data preprocessing.}
At each client, unstructured PDF/HTML sources are processed locally into an instruction-tuning format via a lightweight pipeline: \emph{extract} $\rightarrow$ \emph{clean} $\rightarrow$ \emph{segment} $\rightarrow$ \emph{FAQ-to-QA pairs}. Table~\ref{tab:dataset_stats} reports dataset statistics; the larger SIDFORS documentation (31 pages vs.\ 8 pages) naturally induces a realistic non-IID imbalance.

\begin{table}[h]
\centering
\caption{Dataset statistics per federated client.}
\label{tab:dataset_stats}
\footnotesize
\setlength{\tabcolsep}{2pt}
\renewcommand{\arraystretch}{1.15}
\begin{tabularx}{\columnwidth}{@{}p{2.2cm}XX@{}}
\toprule
\textbf{Metric} & \textbf{Client 1 (SIGESON)} & \textbf{Client 2 (SIDFORS)} \\
\midrule
Primary domain & Radio broadcasting & Electronic communications \\
Source docs & Manual v1 (8 pages) + FAQs & Manual v2.1 (31 pages) + FAQs \\
Key procedures & Turnover declaration; accounting-module access & Network declarations; accounting-module access \\
Samples & 32 & 45 \\
Avg.\ input length (tokens) & 10.0 & 10.19 \\
Avg.\ output length (tokens) & 27.29 & 25.13 \\
\bottomrule
\end{tabularx}
\end{table}

All documents are written in Italian and reflect formal administrative language and platform-specific terminology.

\subsection{Roles and Permissions}
In PA environments, users interact with digital services based on defined institutional roles, which regulate access to information and procedures. GuidaPA uses RBAC to ensure context-appropriate responses and prevent disclosure beyond a user’s permissions. In our experimental setup, each domain’s public documentation is isolated to validate access control: SIDFORS roles receive answers derived only from SIDFORS content, while SIGESON roles receive answers derived only from SIGESON content.

Accordingly, the following roles were defined:
\begin{itemize}
\item \textbf{Radio broadcasting operator/citizen}: external user querying GuidaPA for SIGESON public documentation (manuals and FAQs).
\item \textbf{Public electronic communications operator/citizen}: external user querying GuidaPA for SIDFORS public documentation (manuals and FAQs).
\item \textbf{SIDFORS Officer}: MIMIT officer responsible for managing SIDFORS procedures.
\item \textbf{SIGESON Officer}: MIMIT officer responsible for managing SIGESON procedures.
\item \textbf{Master}: highest-level profile with access to all sources available in GuidaPA.
\end{itemize}

The relationships between roles and access permissions are summarized in Table~\ref{tab:access}.

\begin{table}[h]
\centering
\caption{Access permissions across public documentation.}
\label{tab:access}
\footnotesize
\setlength{\tabcolsep}{4pt}
\renewcommand{\arraystretch}{1.1}
\begin{tabular}{@{}lcc@{}}
\toprule
\textbf{Role} & \textbf{SIDFORS} & \textbf{SIGESON} \\
\midrule
Radio broadcasting operator/citizen         & No & Yes \\
Public electronic comm.\ operator/citizen   & Yes & No \\
SIDFORS officer                     & Yes & No \\
SIGESON officer                     & No & Yes \\
Master                              & Yes & Yes \\
\bottomrule
\end{tabular}
\end{table}

% %%%%%%%%%%%%%%%%%%%%%%%%%%%%%%%%%%%%%%%%%
% %%%%%%%%%%%%%%%%%%%%%%%%%%%%%%%%%%%%%%%%%
\begin{figure*}[t]
\centering
\includegraphics[width=0.9\textwidth]{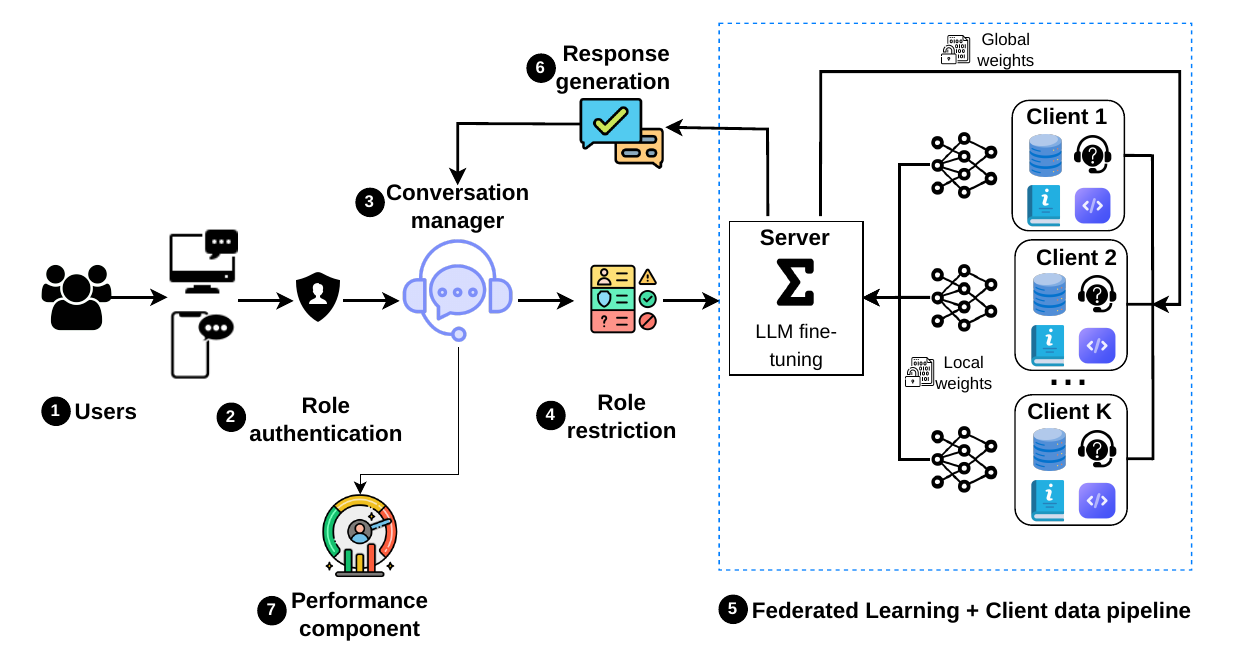}
\caption{The GuidaPA proposed architecture.}
\label{fig:GuidaPAarchitecture}
\end{figure*}
\section{The GuidaPA System} \label{sec:system}

The overall architecture of GuidaPA, depicted in Figure~\ref{fig:GuidaPAarchitecture}, combines natural-language interaction with strong privacy guarantees through FL. It integrates four components---the Conversation Manager, the FL module, the (private) Client Data Pipeline, and a Performance Module---to enable operation across distributed PA data sources. Each PA office acts as a client that trains locally on its own data, while a central server orchestrates aggregation of model updates in a client--server FL workflow. We assume an honest-but-curious server that only aggregates updates; clients keep raw documents local; RBAC is enforced at the Conversation Manager. This setup mirrors the organizational reality of the PA, where data are siloed across offices while processes and regulations remain shared.

\subsection{Conversation Manager Module}
The Conversation Manager is the entry point for user interaction and governs dialogue flow. It provides two safeguards: \emph{role restriction} and \emph{controlled response delivery}. When a user submits a query, the module validates credentials and maps the request to the sources authorized for that role. As shown in Figure~\ref{fig:GuidaPAarchitecture}, role restriction augments the query with the allowed sources while blocking access to restricted content (e.g., via prompt injection~\cite{liu2024formalizing}), thus enforcing least privilege and organizational access policies. The resulting query is sent to the current global model, and the generated answer can be post-processed for conciseness, provenance logging, and safety checks before being returned.

\subsection{FL Module}
At the core of GuidaPA lies the FL module, which enables privacy-preserving collaborative training across distributed PA data sources. Each participating client trains a local model on its own dataset and periodically shares model updates---rather than raw data---with the central server. The server aggregates these updates into a global model, which is redistributed back to clients over multiple rounds. This enables the global model to benefit from collective knowledge while sensitive information remains local, reducing the need for centralized data collection and supporting privacy and regulatory constraints (e.g., GDPR principles).

In this work, we do not train LLMs from scratch. Instead, we perform parameter-efficient fine-tuning of pretrained LLMs, which reduces computational cost and training time while adapting models to the PA domain.

\subsection{Client Data Pipeline Module}
The (private) Client Data Pipeline manages secure preparation and processing of local datasets before training or inference. Each client performs local text extraction, cleaning, and instruction-format conversion, ensuring sensitive data never leaves the client environment. The pipeline is extensible to additional sources, but in this study we use only public manuals and FAQs.

\subsection{Performance Module} \label{sec:performance_module}
This module evaluates the effectiveness of the fine-tuned LLMs within the GuidaPA architecture. It reports quantitative metrics capturing the quality and reliability of generated answers after federated fine-tuning (see Section~\ref{sec:experiments}), enabling monitoring of model behavior across federated rounds.

In addition to answer quality, the Performance Module measures heterogeneity across clients (non-IID data), which is relevant in PA settings where platforms contribute related but distinct procedures and terminology. We quantify non-IID effects using standard distributional measures such as Jensen--Shannon divergence (JSD) and Hellinger distance (HD)~\cite{gutierrez2024fedartml}. Joint analysis of quality metrics and heterogeneity supports assessing the robustness of federated fine-tuning under realistic data imbalance and domain shifts.

\subsection{Chatbot Interface}
\begin{figure*}[t]
\centering
\includegraphics[width=\textwidth]{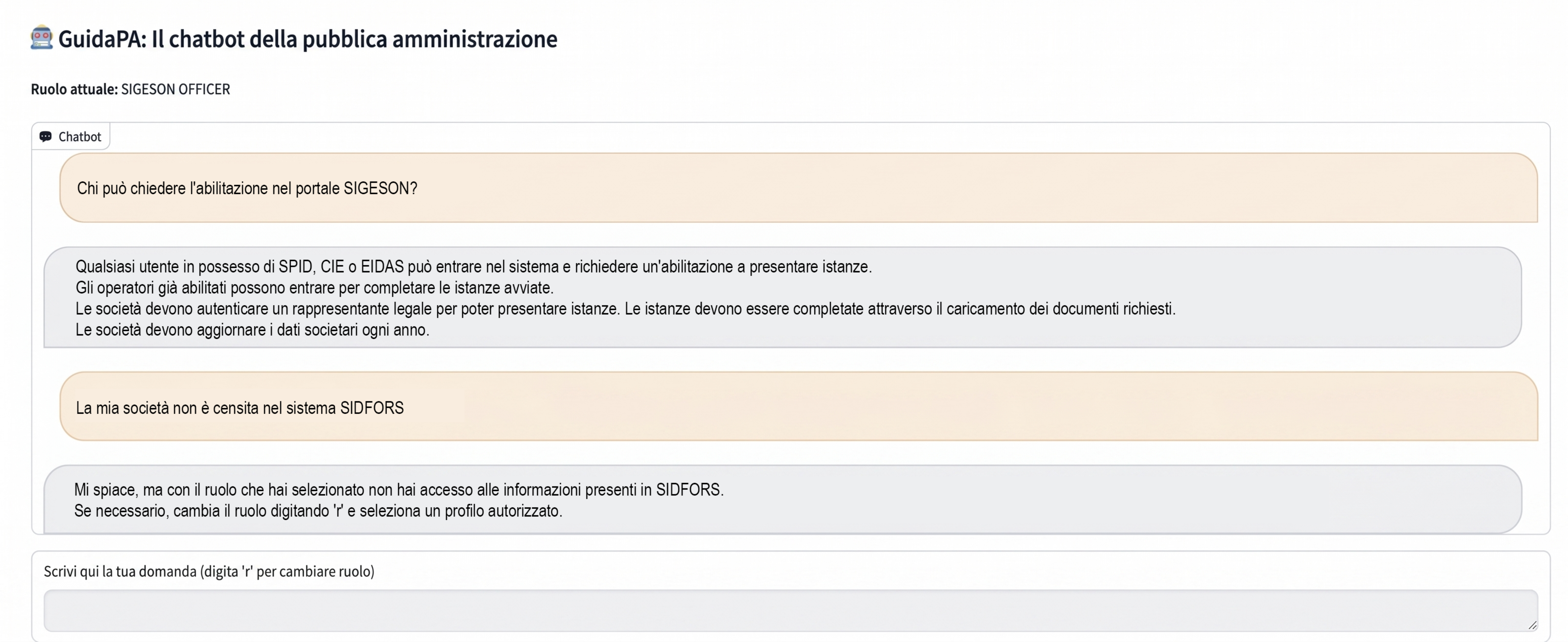}
\caption{The GuidaPA user interface.}
\label{fig:screenshot_chatbot}
\end{figure*}

Figure~\ref{fig:screenshot_chatbot} shows the GuidaPA graphical interface. The active role (e.g., \textbf{SIGESON OFFICER}) determines which information sources are accessible. The interface illustrates two example queries and the corresponding behavior.

In the first query, \emph{``Chi può richiedere l'abilitazione nel portale SIGESON?''} (``Who can request access authorization in the SIGESON portal?''), the system identifies the request as SIGESON-related and returns an answer consistent with the role permissions.

In the second query, \emph{``La mia società non è censita nel sistema SIDFORS''} (``My company is not registered in the SIDFORS system''), the request is recognized as SIDFORS-related. However, as defined by the role-based access policies (Table~\ref{tab:access}), SIGESON officers cannot access SIDFORS resources; the system therefore returns a denial message and prompts the user to change role.

This example highlights how the interface enforces \emph{role-based access restrictions}, ensuring that each profile can query only the sources it is institutionally authorized to consult, consistent with the least-privilege principle.

% %%%%%%%%%%%%%%%%%%%%%%%%%%%%%%%%%%%%%%%%%
% %%%%%%%%%%%%%%%%%%%%%%%%%%%%%%%%%%%%%%%%%
\section{Experiments} \label{sec:experiments}

This section evaluates GuidaPA by assessing whether federated fine-tuning can effectively adapt LLMs to PA documentation under realistic data locality constraints. We describe the experimental setup (clients, training protocol, models, and metrics) and report global and local performance under FL.

\subsection{Experimental Design}\label{ED}

\textbf{Testbed.}
Experiments were run on a single GPU workstation (RTX A6000, 48GB) using Flower~\cite{beutel2020flower} and Python~3.10.

\textbf{Data and Clients.}
Federated experiments use the two PA clients described in Section~\ref{sec:data-sources} (SIGESON and SIDFORS), each training locally on its own documentation. We adopt an 80/20 train--test split per client. The federation runs for 15 server rounds with one local epoch per round (Table~\ref{tab:llm-finetune-hparams}).

\begin{figure}[h]
\centering
\includegraphics[width=\columnwidth]{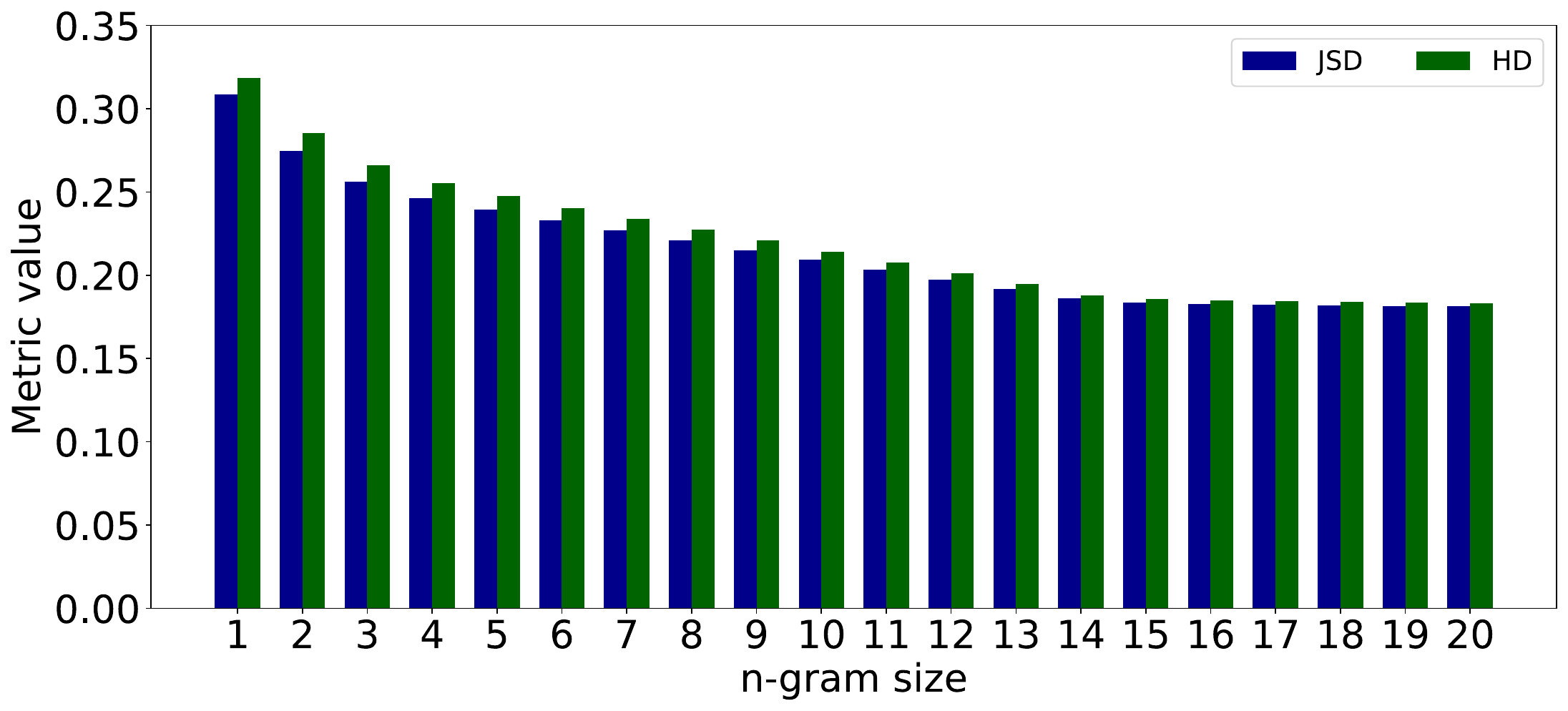}
\caption{Lexical shift (non-IID quantification) between SIGESON and SIDFORS using HD and JSD for multiple $n$-grams.}
\label{fig:lexical_shift__JSD_HD_vs_ngram}
\end{figure}

\textbf{Non-IID Data Quantification.}\label{IID}
To characterize client heterogeneity, we quantify lexical shift between SIGESON and SIDFORS using Hellinger distance (HD) and Jensen--Shannon divergence (JSD) over different $n$-gram sizes. Figure~\ref{fig:lexical_shift__JSD_HD_vs_ngram} shows moderate divergence, with higher differences for smaller $n$-grams and increasing overlap for longer $n$-grams, consistent with shared administrative language but distinct domain terminology. These measurements support analyzing FL robustness under realistic non-IID conditions (Section~\ref{sec:performance_module})~\cite{gutierrez2024fedartml}.

\textbf{Models and Baselines.}
We evaluate a focused set of Italian-centric and multilingual instruction-tuned LLMs commonly used in Italian NLP benchmarks~\cite{moroni2024towards,orlando2024minerva} and available via the Transformers ecosystem~\cite{wolf-etal-2020-transformers}. Specifically, we consider Italian-oriented conversational models and strong multilingual baselines: \texttt{galatolo/cerbero-7b} and \texttt{galatolo/cerbero-7b-openchat}; \texttt{cosimoiaia/Loquace-7B-Mistral}; \texttt{mii-llm/\allowbreak maestrale-chat-v0.4-beta}; \texttt{giux78/ \allowbreak zefiro-7b-beta-ITA-v0.1}; Meta’s multilingual \texttt{meta-llama/Llama-3.1-8B-Instruct}; and finally the Mistral family (\texttt{mistralai/ Mistral-7B-Instruct-v0.3} and \texttt{mistralai/\allowbreak Mistral-7B-v0.1}). All models are adapted with the same federated parameter-efficient setup (Table~\ref{tab:llm-finetune-hparams}) using QLoRA and low-rank adaptation, following established best practices for efficient fine-tuning in resource- and communication-constrained settings~\cite{hu2022lora,dettmers2023qlora,wang2024flora,gao2025flowertune}. We use FedAvg with weighted averaging by client dataset size.

\begin{table}[h]
\centering
\caption{Hyperparameters used for LLM fine-tuning in FL.}
\label{tab:llm-finetune-hparams}
\footnotesize
\setlength{\tabcolsep}{3pt}
\renewcommand{\arraystretch}{1.15}
\begin{tabular}{ll}
\toprule
\textbf{Hyperparameter} & \textbf{Value} \\
\midrule
Parameter-efficient method & QLoRA \\
Quantization (bits) & 4 \\
Gradient checkpointing & true \\
LoRA rank ($r$) & 32 \\
LoRA $\alpha$ & 64 \\
LoRA target modules & q\_proj, k\_proj, v\_proj, o\_proj \\
Sequence length & 512 \\
Per-device batch size & 4 \\
Gradient accumulation & 4 steps (effective batch $=16$) \\
Learning rate & $1\times10^{-4}$ \\
LR scheduler & cosine \\
Max training steps & 100 \\
Num train epochs & 1 \\
Logging steps & 50 \\
Save steps / limit & 1000 / 10 \\
Local epochs (\# client epochs) & 1 \\
Rounds (\# server rounds) & 15  \\
\bottomrule
\end{tabular}
\end{table}

Although the dataset size is intentionally small—serving as a safe, public proxy for restricted institutional sources —we actively mitigate the risk of overfitting through our architectural and hyperparameter choices. First, the use of PEFT via QLoRA restricts the trainable weights to a small set of low-rank adapters (with \texttt{r=32}). This significantly reduces the model's capacity to memorize the training data, acting as a strong regularizer compared to full fine-tuning. Second, the federated training configuration strictly limits local adaptation to a single epoch per communication round across a total of 15 rounds. This strategy prevents the local models from overfitting their limited respective samples, while the central Server's periodic aggregation enforces a regularizing consensus across the clients.

\textbf{Metrics.}\label{subsec:metrics}
We evaluate answer quality using standard Natural Language Generation (NLG) metrics~\cite{hu2404unveiling,banerjee2308benchmarking}, covering recall-oriented overlap, precision-oriented overlap, and semantic similarity:
\begin{itemize}
    \item \textbf{ROUGE-1/2/L}: unigram overlap, bigram overlap, and longest common subsequence (recall-oriented).
    \item \textbf{BLEU-4}: 4-gram precision overlap, rewarding fluency while penalizing omissions/extraneous content.
    \item \textbf{METEOR}: synonym/stem-aware matching with word-order sensitivity, capturing semantic adequacy.
\end{itemize}

% %%%%%%%%%%%%%%%%%%%%%%%%%%%%%%%%%%%%%%%%%
% %%%%%%%%%%%%%%%%%%%%%%%%%%%%%%%%%%%%%%%%%
\section{Results} \label{sec:results}

Before analyzing the federated setup, we establish a reference point using the Private Centralized model as the ideal case where all client data is available in a single secure environment. This baseline allows comparison between the federated and centralized settings. We report (i) \emph{local} metrics on each client’s test set and (ii) \emph{global} FL metrics as a dataset-size-weighted average across clients. Together, ROUGE measures content coverage, BLEU-4 exactness and fluency, and METEOR semantic similarity and linguistic flexibility.

\subsection{The baseline: the Private Centralized Model}\label{baseline}

To address the issue of lack of domain-specific understanding highlighted in Section~\ref{sec:motiv}, we first establish a baseline using a Private Centralized Model (see Figure~\ref{fig:three_architectures}). This configuration represents a conventional training scenario in which all available training data (public docs in this study) from SIGESON and SIDFORS are combined within a single, secure environment under PA control. The goal is to assess how much fine-tuning a general-purpose LLM on domain-specific documentation improves its ability to generate accurate and context-aware responses. This baseline serves as a reference point for evaluating the benefits and potential trade-offs of the federated approach introduced later.

\begin{table}[h]
\centering
\caption{Centralized evaluation metrics for general-purpose and fine-tuned (on SIDFORS and SIGESON data) approaches. The model used is the one that yields the best results in the federated global results (see Table~\ref{tab:metric-global}). In bold, the highest value of each metric.}
\label{tab:metric-centralized}
\resizebox{\columnwidth}{!}{
\begin{tabular}{lrrrrr}
\toprule
\textbf{Model} & \textbf{ROUGE-1} & \textbf{ROUGE-2} & \textbf{ROUGE-L} & \textbf{BLEU-4} & \textbf{METEOR} \\
\midrule
\makecell{\texttt{cerbero-7b-openchat} \\ (general-purpose)} & 41.45 & 29.97 & 38.41 & 26.97 & 41.83 \\
\makecell{\texttt{cerbero-7b-openchat} \\ (fine-tuned)} & \textbf{62.18} & \textbf{55.37} & \textbf{60.09} & \textbf{50.90} & \textbf{64.17} \\
\bottomrule
\end{tabular}
}
\end{table}

Table~\ref{tab:metric-centralized} reports the centralized evaluation metrics obtained with the best-performing model in the federated experiments (see Table~\ref{tab:metric-global}). The results show a substantial improvement across all metrics when the model is fine-tuned on PA-specific data (SIGESON and SIDFORS) compared to its general-purpose version. ROUGE and METEOR scores increase by more than 20 points on average, while BLEU-4 rises by nearly 24 points, confirming that fine-tuning allows the model to better capture the specialized terminology and procedural language of the PA domain.

\subsection{FL Model}

This section presents the evaluation of the FL model implemented in GuidaPA (see Figure~\ref{fig:three_architectures}), first analyzing its performance vs.\ that of the private centralized baseline, and then considering the performance of the local models running on the clients of the federated architecture.

\begin{table}[h]
\centering
\caption{Global evaluation metrics across all clients with FL models. In bold, the highest value of each metric.}
\label{tab:metric-global}
\resizebox{\columnwidth}{!}{
\begin{tabular}{lrrrrr}
\toprule
\textbf{Model} & \textbf{ROUGE-1} & \textbf{ROUGE-2} & \textbf{ROUGE-L} & \textbf{BLEU-4} & \textbf{METEOR} \\
\midrule
\texttt{cerbero-7b} & \textbf{61.14} & 53.72 & 59.09 & 44.30 & 62.99 \\
\texttt{cerbero-7b-openchat} & 61.10 & \textbf{55.77} & \textbf{59.44} & \textbf{45.02} & \textbf{63.94} \\
\texttt{Llama-3.1-8B-Instruct} & 45.86 & 38.45 & 44.83 & 29.32 & 50.83 \\
% \texttt{LLaMAntino-3-ANITA-8B-Inst} & 50.96 & 43.84 & 48.63 & 33.54 & 56.93 \\
\texttt{Loquace-7B-Mistral} & 53.10 & 46.81 & 51.68 & 37.63 & 54.75 \\
% \texttt{maestrale-chat-v0.1-alpha-sft} & 60.27 & 53.82 & 58.39 & 42.75 & 61.62 \\
\texttt{maestrale-chat-v0.4-beta} & 49.25 & 40.90 & 46.82 & 32.26 & 50.05 \\
% \texttt{Minerva-3B-base-v1.0} & 30.39 & 19.23 & 27.49 & 13.38 & 31.98 \\
% \texttt{Minerva-7B-instruct-v1.0} & 46.04 & 38.58 & 43.59 & 27.73 & 56.52 \\
\texttt{Mistral-7B-Instruct-v0.3} & 53.02 & 43.26 & 49.91 & 34.33 & 54.27 \\
\texttt{Mistral-7B-v0.1} & 50.30 & 42.38 & 47.63 & 34.53 & 53.71 \\
% \texttt{OLMo-7B-0724-Instruct-hf} & 49.82 & 41.16 & 48.66 & 32.57 & 51.03 \\
\texttt{zefiro-7b-beta-ITA-v0.1} & 58.55 & 50.97 & 56.30 & 39.43 & 61.57 \\
\bottomrule
\end{tabular}
}
\end{table}

\subsubsection{Comparison with the Private Centralized Baseline} \label{sec:comparison}

\begin{table*}[t]
\centering
\caption{Local evaluation metrics for client 1 (SIGESON) and client 2 (SIDFORS). In bold, the highest value of each metric.}
\label{tab:metric-local}
\resizebox{\textwidth}{!}{
\begin{tabular}{lrrrrr|rrrrr}
\toprule
 & \multicolumn{5}{c}{\textbf{SIGESON}} & \multicolumn{5}{c}{\textbf{SIDFORS}} \\
\cmidrule(lr){2-6}\cmidrule(lr){7-11}
\textbf{Model} & ROUGE-1 & ROUGE-2 & ROUGE-L & BLEU-4 & METEOR & ROUGE-1 & ROUGE-2 & ROUGE-L & BLEU-4 & METEOR \\
\midrule
\texttt{cerbero-7b} & \textbf{68.90} & \textbf{62.57} & \textbf{67.16} & \textbf{49.61} & \textbf{67.79} & 52.79 & 44.19 & 50.41 & 38.59 & 57.81 \\
\texttt{cerbero-7b-openchat} & 62.64 & 57.54 & 60.99 & 44.73 & 64.67 & \textbf{59.45} & \textbf{53.87} & \textbf{57.78} & \textbf{45.33} & \textbf{63.17} \\
\texttt{Llama-3.1-8B-Instruct} & 53.47 & 46.13 & 51.64 & 31.61 & 56.96 & 37.66 & 30.19 & 37.50 & 26.85 & 44.24 \\
% \texttt{LLaMAntino-3-ANITA-8B-Inst} & 56.58 & 51.62 & 55.00 & 38.34 & 61.95 & 44.91 & 35.47 & 41.77 & 28.37 & 51.52 \\
\texttt{Loquace-7B-Mistral} & 58.91 & 52.64 & 56.86 & 41.51 & 59.23 & 46.84 & 40.53 & 46.11 & 33.45 & 49.93 \\
% \texttt{maestrale-chat-v0.1-alpha-sft} & 64.03 & 59.04 & 62.40 & 45.77 & 64.31 & 56.22 & 48.20 & 54.07 & 39.51 & 58.71 \\
\texttt{maestrale-chat-v0.4-beta} & 60.26 & 53.29 & 57.70 & 39.73 & 59.47 & 37.39 & 27.55 & 35.09 & 24.22 & 39.91 \\
% \texttt{Minerva-3B-base-v1.0} & 33.20 & 21.44 & 30.20 & 15.77 & 33.97 & 27.36 & 16.85 & 24.57 & 10.81 & 29.84 \\
% \texttt{Minerva-7B-instruct-v1.0} & 50.54 & 42.89 & 47.75 & 29.44 & 61.53 & 41.19 & 33.94 & 39.12 & 25.88 & 51.11 \\
\texttt{Mistral-7B-Instruct-v0.3} & 52.63 & 40.67 & 47.43 & 29.92 & 50.12 & 53.45 & 46.05 & 52.59 & 39.07 & 58.74 \\
\texttt{Mistral-7B-v0.1} & 57.39 & 49.89 & 53.92 & 39.64 & 59.28 & 42.67 & 34.30 & 40.86 & 29.03 & 47.72 \\
% \texttt{OLMo-7B-0724-Instruct-hf} & 50.77 & 41.83 & 48.39 & 32.24 & 51.54 & 48.79 & 40.42 & 48.95 & 32.93 & 50.49 \\
\texttt{zefiro-7b-beta-ITA-v0.1} & 67.09 & 60.45 & 64.43 & 46.46 & 65.95 & 49.34 & 40.75 & 47.54 & 31.85 & 56.84 \\
\bottomrule
\end{tabular}
}
\end{table*}

We compare \texttt{cerbero-7b-openchat}, the best global federated model (Table~\ref{tab:metric-global}), against its centralized counterpart (Table~\ref{tab:metric-centralized}). The comparison shows that federated metrics are close to those achieved under centralized training, indicating that federated fine-tuning can preserve model quality while keeping data local and maintaining data sovereignty.

\subsubsection{FL Model Performance Evaluation} \label{sec:global_performance}

We report final-round values for comparison. Table~\ref{tab:metric-global} reports the final global metric values obtained at the last communication round for the eight models considered. Overall, Italian instruction-tuned models yield the strongest trade-off across recall-, precision-, and semantic-oriented metrics. These trends align with prior findings that instruction-tuned Italian models are strong on generation-based evaluation, but here we validate them in a PA deployment setting~\cite{ragazzi2024lawsuit,sarti2022it5,priola2024addressing}.

\subsubsection{Local Model Performance Evaluation}

Table~\ref{tab:metric-local} reports local evaluation metrics for each client (SIGESON and SIDFORS). Overall, most models achieve higher ROUGE, BLEU-4, and METEOR on SIGESON, suggesting that SIGESON is more favorable for adaptation, whereas SIDFORS is more challenging.

On SIGESON, \texttt{cerbero-7b} achieves the best results across all metrics (e.g., ROUGE-L = 67.16, BLEU-4 = 49.61, METEOR = 67.79), with \texttt{zefiro-7b-beta-ITA-v0.1} also performing strongly (ROUGE-L = 64.43, METEOR = 65.95). On SIDFORS, \texttt{cerbero-7b-openchat} is the top model, leading all five metrics (ROUGE-1 = 59.45, ROUGE-2 = 53.87, ROUGE-L = 57.78, BLEU-4 = 45.33, METEOR = 63.17), while \texttt{Mistral-7B-Instruct-v0.3} is the strongest remaining baseline among those reported (e.g., ROUGE-L = 52.59, METEOR = 58.74). These local differences highlight client non-IID data, while the global aggregation (Table~\ref{tab:metric-global}) supports a robust overall model.

% %%%%%%%%%%%%%%%%%%%%%%%%%%%%%%%%%%%%%%%%%
% %%%%%%%%%%%%%%%%%%%%%%%%%%%%%%%%%%%%%%%%%
\section{Discussion and Implications}
\label{sec:discussion}

Our results show that federated, parameter-efficient adaptation of LLMs can support privacy-preserving conversational AI in PA while retaining strong answer quality. This is relevant because federated LLM deployment is often constrained by client heterogeneity, non-IID data, communication/efficiency limits in PEFT, and application-level security concerns~\cite{yao2024federated,piccialli2025federated,liu2024formalizing}. In the PA setting, these challenges are further shaped by organizational silos and governance requirements.

\subsection{Why it matters for PA}
\textbf{Near-centralized quality without centralizing data.}
Across ROUGE, BLEU-4, and METEOR, federated fine-tuning achieves performance close to private centralized training, indicating that PA entities can collaborate to build high-quality assistants while keeping documentation local. This provides a practical alternative to building centralized data repositories when data sharing is slow, expensive, or constrained by governance.

\textbf{Moderate non-IID supports stable aggregation in institutional domains.}
By quantifying lexical divergence and relating it to observed behavior (Section~\ref{sec:experiments}), we find that cross-client differences remain moderate, consistent with PA documentation being produced under shared regulatory and procedural conventions. This structured non-IID setting appears conducive to effective global aggregation, reducing the need for heavy personalization mechanisms often required in highly divergent domains.

\textbf{Security and governance at the application layer.}
LLM-based systems are vulnerable to prompt injection and instruction-following attacks that can bypass intended policies if authorization is not explicitly enforced~\cite{liu2024formalizing}. GuidaPA integrates RBAC and source restriction so different user categories (citizens, operators, officers) only access authorized information. Combined with monitoring of answer quality and non-IID divergence, it supports a governance workflow for deployment and maintenance under evolving local data~\cite{yao2024federated,piccialli2025federated}.

\subsection{How this study differs from existing work}
To clarify novelty, GuidaPA differs from prior research in the following ways:
\begin{itemize}
    \item \textbf{Domain novelty:} This is the first federated LLM-based chatbot study grounded in an operational Italian PA context, rather than healthcare, education, or social platforms.
    \item \textbf{End-to-end deployment focus:} Beyond model training, we integrate RBAC, client-side secure preprocessing, and monitoring of both quality and non-IID data as first-class components.
    \item \textbf{Empirical evidence under realistic PA non-IID data:} We quantify cross-client divergence and show that federated tuning remains stable and near-centralized in this institutional domain, complementing recent federated PEFT work that targets extreme non-IID task regimes~\cite{wang2024flora,bai2024federated}.
\end{itemize}

Overall, GuidaPA serves both as empirical evidence that federated PEFT can be effective under realistic institutional constraints and as a practical blueprint for privacy-preserving conversational AI in information-intensive public services.

% %%%%%%%%%%%%%%%%%%%%%%%%%%%%%%%%%%%%%%%%%
% %%%%%%%%%%%%%%%%%%%%%%%%%%%%%%%%%%%%%%%%%
\section{Limitations}
\label{sec:limitations}

This study has the following limitations.
First, our empirical evaluation involves two PA platforms (SIGESON and SIDFORS) and uses public manuals/FAQs as a safe proxy for restricted institutional sources; consequently, results primarily reflect adaptation on small, authoritative documentation under moderate non-IID data.
In particular, the amount of domain material used for adaptation is limited (approximately 8 and 31 pages, respectively), which makes this setting more manageable than realistic deployments involving larger, noisier, and more heterogeneous institutional corpora. Therefore, the results should be interpreted mainly as evidence of feasibility rather than as evidence that the proposed framework will scale unchanged to broader PA settings.
Second, evaluation is based on offline text-generation metrics (ROUGE, BLEU-4, METEOR) on held-out splits and does not include human assessment~\cite{chaudhari2024rlhf}, end-to-end task success, or live deployment measurements.
Third, the federated training uses a standard aggregation baseline and does not compare alternative optimization strategies specifically designed to improve robustness under stronger non-IID conditions (e.g., proximal objectives or momentum-based server updates).

% %%%%%%%%%%%%%%%%%%%%%%%%%%%%%%%%%%%%%%%%%
% %%%%%%%%%%%%%%%%%%%%%%%%%%%%%%%%%%%%%%%%%
\section{Conclusion and Future Work} \label{sec:conclusion}

In conclusion, GuidaPA demonstrates the feasibility of a privacy-preserving chatbot for PA through federated fine-tuning of LLMs on siloed institutional data. The system combines RBAC, performance monitoring, and non-IID data quantification to produce accurate, context-aware, and policy-compliant answers. Results show that, despite moderate non-IID data across clients, the federated setup yields robust models for both citizens and officers, with performance comparable to centralized models. Overall, this work shows that FL can balance scalability, data protection, and usability in AI-driven public services.

Future work will extend GuidaPA in three directions.
(i) \emph{Broader data and clients:} incorporate additional PA entities and restricted sources (e.g., ticketing archives, officer manuals, curated database extracts) to study stronger non-IID settings, evolving documentation, and multi-agency governance.
(ii) \emph{Non-IID robustness and optimization:} evaluate federated optimizers for non-IID clients---such as FedProx~\cite{li2020federated} and momentum-based aggregation (e.g., FedAvgM~\cite{hsu2019measuring})---and assess their impact on convergence and local/global quality under more diverse client distributions.
(iii) \emph{Stronger privacy/security and evaluation:} integrate secure aggregation and/or differential privacy, expand authorization testing via systematic red-teaming, and complement automatic metrics with human and task-based evaluation.

\section*{Acknowledgments}
Daniel M.\ Jimenez-Gutierrez was partially supported by PNRR351
TECHNOPOLE -- NEXT GEN EU Roma Technopole -- Digital Transition,
FP2 -- Energy transition and digital transition in urban regeneration and
construction. 
Andrea Vitaletti was supported by the project SERICS
(PE00000014) under the MUR National Recovery and Resilience Plan
funded by the European Union - NextGenerationEU.

\bibliographystyle{IEEEtran}
\bibliography{00_references}

\end{document}